\documentclass{article}

\usepackage{arxiv}

\usepackage[utf8]{inputenc} 
\usepackage[T1]{fontenc}    
\usepackage[colorlinks,bookmarksopen,bookmarksnumbered,citecolor=red,urlcolor=red]{hyperref}       
\usepackage{booktabs}       
\usepackage{amsmath,amsfonts}       
\usepackage{nicefrac}       
\usepackage{microtype}      
\usepackage{lipsum}		
\usepackage{graphicx}
\usepackage{natbib}
\usepackage{doi}

\usepackage{algorithmic}
\usepackage{amssymb}
\usepackage[caption=false,font=normalsize,labelfont=sf,textfont=sf]{subfig}
\usepackage{graphicx}
\usepackage{balance}
\usepackage{moreverb}
\usepackage{calc}
\usepackage{accents}
\usepackage{tabularray}
\usepackage[shortlabels]{enumitem}
\usepackage{xcolor}
\usepackage{adjustbox}

\title{Abstractive Summarization of\\Large Document Collections Using GPT}

\date{June 11, 2023}	

\author{{Shengjie Liu}\\
	Operations Research Department\\
	North Carolina State University\\
	Raleigh, NC 27695-8206 \\
	\texttt{sliu52@ncsu.edu} \\
	\And
	\href{https://orcid.org/0000-0002-2617-8638}{\includegraphics[scale=0.06]{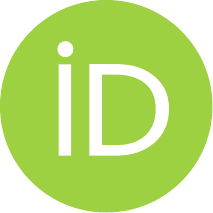}\hspace{1mm}Christopher G.~Healey} \\
	Department of Computer Science \& Institute for Advanced Analytics\\
	North Carolina State University\\
	Raleigh, NC 27695-8206 \\
	\texttt{healey@ncsu.edu} \\
}

\renewcommand{\headeright}{}
\renewcommand{\undertitle}{}
\renewcommand{\shorttitle}{\textit Abstractive Summarization of Large Document Collections}

\hypersetup{
pdftitle={A template for the arxiv style},
pdfsubject={q-bio.NC, q-bio.QM},
pdfauthor={David S.~Hippocampus, Elias D.~Striatum},
pdfkeywords={First keyword, Second keyword, More},
}

\begin{document}
\maketitle

\begin{abstract}
This paper proposes a method of abstractive summarization designed to scale to document collections instead of individual documents. Our approach applies a combination of semantic clustering, document size reduction within topic clusters, semantic chunking of a cluster's documents, GPT-based summarization and concatenation, and a combined sentiment and text visualization of each topic to support exploratory data analysis. Statistical comparison of our results to existing state-of-the-art systems BART, BRIO, PEGASUS, and MoCa using ROGUE summary scores showed statistically equivalent performance with BART and PEGASUS on the CNN/Daily Mail test dataset, and with BART on the Gigaword test dataset. This finding is promising since we view document collection summarization as more challenging than individual document summarization. We conclude with a discussion of how issues of scale are being addressed in the GPT large language model, then suggest potential areas for future work.
\end{abstract}


\noindent
\textbf{\textit{ACM-Class}}\\[-18pt]
\begin{itemize}
\item \textbf{Computing methodologies $\sim$ Artificial intelligence $\sim$ Natural language processing $\sim$ Natural Language Generation}
\item \textbf{Information systems $\sim$ Information retrieval $\sim$ Retrieval tasks and goals $\sim$ Summarization}
\item \textbf{Theory of computation $\sim$ Design and analysis of algorithms $\sim$ Online algorithms}
\item \textbf{Human-centered computing $\sim$ Visualization $\sim$ Visualization application domains $\sim$ Visual analytics}
\end{itemize}

\section{Introduction}

Research on transformer--attention mechanisms and large language models (LLMs) have produced impressive results, particularly in natural language processing (NLP) and text analytics. LLMs like BERT \cite{kenton:2019}, BART \cite{lew:2020}, GPT \cite{rad:2018}, Bard \cite{man:2023}, and LLaMA \cite{meta:2023} have produced significant research and general public impact. Despite their state-of-the-art performance, the goal of broad, general-purpose use leaves certain tasks only partially solved. This paper focuses on the abstractive summarization of multi-document collections. Systems like GPT can perform abstractive summarization but are currently limited to a small maximum input of 512 to 4,096 terms. Document collections can easily consist of hundreds of documents containing thousands of terms. An intelligent method is needed to manage scale to leverage an LLM's abstractive summarization capabilities. We also propose applying sentiment analysis and visualization to augment the summaries with additional properties presented in an interactive and simple-to-understand visual format.

Our approach performs the following steps to extend GPT's abstractive summarization method to large document collections.

\begin{enumerate}

\item Apply the Facebook AI Similarity Search \cite{johnson2017billionscale} (FAISS) to estimate document similarity based on the semantic similarity of pairs of documents.

\item Perform Hierarchical Density-Based Spatial Clustering and Application with Noise \cite{Malzer_2020} (HDBSCAN) using FAISS results to generate semantic topic clusters.

\item Identify topic-representative terms and build a collection of \textit{representative term sets} for each cluster, where each set contains a representative term and all semantically similar terms \cite{nagwani2015summarizing} in the parent cluster.

\item Use the representative term sets to further reduce topic cluster size by combining sentences in a cluster containing representative terms into \textit{semantic chunks} based on change points in their semantic content.

\item Use GPT's summarization API to summarize each semantic chunk, followed by its concatenation API to combine the semantic chunk summaries into an abstractive summarization of the original document collection.

\item Perform term-based sentiment analysis on each semantic chunk to generate valence (pleasure) and arousal scores.

\item Visualize the semantic chunks in a dashboard that allows interactive exploration of both the summary's sentiment and its text at different levels of detail.

\end{enumerate}

Comparison of our summaries with current state-of-the-art approaches using ROGUE metrics shows we achieve comparable performance for a multi-document collection versus single document--summary pairs. This suggests our approach can effectively scale to summarize larger documents or document collections beyond the scope of existing systems.

\section{Related Work}

Text summarization is generally divided into two broad categories: \textit{extractive} and \textit{abstractive}. Several recent survey papers cover this topic in detail \cite{cao:2022,gup:2019,lin:2019,mor:2016,zha:2022b}.

\subsection{Summarization}

Extractive summarization extracts representative text from the original documents verbatim and organizes it into a coherent summary. Common approaches include keyword extraction using a weighting scheme to rank keywords based on how well they capture a document's content (\textit{e.g.}, term frequency--inverse document frequency) or sentence extraction using weights to rank sentences and measures of sentence--sentence similarity to avoid including redundant text.

Abstractive summarization attempts to construct a unique text summary without using text from a document verbatim. This is similar to how a human reader would construct a summary. Lin and Ng describe this as a three-step process of information extraction, content selection, and surface realization \cite{lin:2019}.

\begin{itemize}

\item \textbf{Information extraction.} Extract meaningful information from a document, for example, noun and verb phrases together with context, information items that form subject-verb-object triples, or verb--object abstraction schemas for different topics.

\item \textbf{Content selection.} Select a subset of candidate phrases to include in the summary. Heuristic approaches and integer linear programming (ILP) have been proposed to structure the summarization task as a constrained optimization problem. The advantage of ILP is that phrase selection is performed \textit{jointly} across all candidates and not \textit{sequentially}.

\item \textbf{Surface realization.} Combine candidate phrases using grammar and syntactic rules to generate a coherent summary.

\end{itemize}

Gupta and Gupta \cite{gup:2019} propose dividing abstractive summarization into \textit{structural}, \textit{semantic}, and \textit{neural network}-based. Structural approaches identify relevant information in the text to create a predefined structure for generating abstractive summaries. Semantic methods convert text into a semantic representation used as input to a natural language generation (NLG) system to create abstractive summaries. Neural network techniques use deep neural networks to train a model using text--summary pairs. The model is then applied to generate abstractive summaries for unlabeled text.

Structure-based methods use trees, templates, ontologies, graphs, and rules to convert text to abstractive summaries. Barzilay and McKeown \cite{Bar:2005} apply a content theme approach to convert common phrases into dependency trees. The dependency trees are augmented with additional relevant information to create a set of subtrees used as input to an NLG system to organize them into novel sentences and produce an abstractive summary. Alternatively, \textit{word graphs} are built to represent term relationships in a document, then analyzed to construct abstractive summarizations \cite{llo:2011}. Mehdad et al. locate the shortest path in a word graph to identify relevant sentences and remove redundant information, then fuse the remaining sentences to produce an abstractive summary \cite{meh:2013}.

Semantic approaches build a semantic representation of the document text, for example, predicate-argument structures (verbs, subjects, and objects of a sentence) or semantic graphs. These are used as input to an NLG system that converts the semantic representation into an abstractive summary. Moawad and Aref built rich semantic graphs where nouns and verbs represent nodes, and semantic relationships represent edges \cite{mow:2012}. The graph is reduced using heuristic rules, then fed to an NLG system. Munot and Govilkar used domain ontologies to connect concepts between sentences before NLG parsing \cite{mun:2015}. Similar approaches have been used with abstract meaning representation graphs: directed acyclic graphs of sentences and their semantic relationships \cite{liu:2015}.

\subsection{Neural Network Abstractive Summarization}

The most recent state-of-the-art abstractive summarizers use deep neural networks to train models based on text--summary pairs. Initial work used recurrent neural networks (RNNs) to generate summaries \cite{buy:2017}. More recent approaches use transformed-based attention models. Both methods are built on encoder--decoder strategies. An encoder converts terms into vector representations, usually word embeddings. A decoder attempts to determine the next word in the output based on the previous words to date. Training an encoder--decoder is often called a \textit{seq2seq} learning problem. Attention allows information to be preferentially extracted from the encoder based on where to focus to gain the largest benefit at the current stage in the training process.

Examples of RNN abstractive summarizers are numerous. Rush et al. construct a feed-forward neural network comprised of an attention-based encoder and a beam-search decoder for sentence-level summarization \cite{cho:2016,rus:2015}. A beam search expands on the traditional greedy search by selecting the $k$ best candidates at each decoding step and pruning them if an end-of-sequence token is seen. Nallapati et al. use an RNN that processes word embeddings and generates summaries to address issues like keyword modeling and rare word inclusion \cite{nal:2016}.

Recent work involves large language models (LLMs) like GPT, Bard \cite{man:2023}, and LLaMA \cite{meta:2023}. GPT-3 (Generative Pre-trained Transformer 3) was built on a training set with 175 billion parameters \cite{bro:2020}. GPT-3 adopts a \textit{meta-learning} approach: constructing a model with a broad set of general skills and pattern recognition abilities during training, then inferring results with at most a few examples during task completion without adjusting any of the model's internal weights. GPT-3 demonstrated performance with zero-shot (no task examples), one-shot (one task example), and few-shot (10--100 task examples) contexts that rivalled fine-tuned models. Tasks included sentence completion, story-ending selection, question answering, language translation, pronoun reference, common sense reasoning, mathematical reasoning, and reading comprehension. Meta's LLaMA (Large Language Model Meta AI) uses an alternative approach, arguing that performance depends not only on parameter size but also on the amount of training performed \cite{tou:2023}. Touvron et al. propose that although a larger model may be less expensive to train, a smaller model trained for more time will be more efficient at \textit{inference} when it is used to perform NLP tasks. LLaMA was evaluated by training it on different-sized inputs, performing a short fine-tuning step, then testing it on question answering, common sense reasoning, mathematical reasoning, code generation, and reading comprehension using 0, 1, 5, and 64-shot contexts.  Training sets contained 7, 13, 33, and 65 billion models, trained on 1 trillion tokens for the two smaller model sets and 1.4 trillion tokens for the two larger model sets. Task performance showed LLaMA outperformed GPT-3 on most benchmarks. 

\subsubsection{Training and Test Datasets.}

A number of datasets containing text and a corresponding summary exist for training and testing.
One of the most common is the Document Understanding Conferences (DUC) datasets managed by the National Institutes of Standards and Technology (NIST) \cite{tac:2011}. Analysis of DUC datasets is a track in NIST's annual Text Analytics Conference. Each DUC entry contains news documents and three ground-truth summaries: (1) manually generated, (2) automatically generated as baselines, and (3) automatically generated by existing systems.
The CNN/Daily Mail dataset contains document--summary pairs of newspaper articles and corresponding summaries: 286,817 training pairs, 13,368 validation pairs, and 11,487 test pairs \cite{see:2023}. The Gigaword dataset contains approximately 3.8 million English news article training pairs, 189,000 validation pairs, and 2,000 testing pairs \cite{gig:2023}. Although sufficient to train a neural network model, the training pairs use the first sentence of a document as its summarization ground truth. Finally, the NYT dataset contains preprocessed articles from the New York Times: approximately 650,000 manually generated article--summary pairs with articles limited to 800 tokens and summaries limited to 100 tokens \cite{san:2008}.

\subsubsection{Evaluation.}
Once a \textit{candidate} abstractive summary has been constructed, it must be compared to a \textit{reference} ``ground truth'' summary to evaluate its quality. Although several evaluation metrics exist, including BLEU, METEOR, and ROUGE, variations of ROUGE (Recall-Oriented Understudy for Gisting Information) are the most common evaluation methods: ROUGE-1 for unigrams, ROUGE-2 for bigrams, and ROUGE-L for longest common subsequence \cite{lin:2004}. ROUGE's evaluations are based on character overlap, not on semantic content. ROUGE returns recall: the proportion of words in the reference summary captured by the candidate summary; and precision: the proportion of words in the candidate summary that appear in the reference summary.

\begin{equation}
\textrm{ROUGE-1}_{\textrm{R}} = \frac{n_{\textrm{o}}}{n_{\textrm{r}}} \hspace*{0.5in}
\textrm{ROUGE-1}_{\textrm{P}} = \frac{n_{\textrm{o}}}{n_{\textrm{c}}}
\end{equation}

\noindent
where $n_{\textrm{c}}$ is the number of candidate tokens, $n_{\textrm{r}}$ is the number of reference tokens, and $n_{\textrm{o}}$ is the number of candidate tokens included in (\textit{i.e.}, overlapping) the reference summary. Consider the reference summary ``John really loves data science'' and the candidate summary ``John loves data science.'' Here $n_{\textrm{r}}=5$, $n_{\textrm{c}}=4$, and $n_{\textrm{o}}=4$ so recall is $\textrm{ROUGE-1}_{\textrm{R}} = \frac{4}{5}=0.8$ and precision is $\textrm{ROUGE-1}_{\textrm{P}} = \frac{4}{4} = 1.0$. ROUGE-2 uses the same formulas for recall and precision but works with bigrams rather than unigrams. For the same candidate and reference sentences $n_{\textrm{r}}=4$: \{(John, really), (really, loves), (loves, data), (data, science)\}; $n_{\textrm{c}}=3$: \{(John, loves), (loves, data), (data, science)\} and $n_{\textrm{o}}=2$: \{(loves, data), (data, science)\}, producing recall and precision of $\textrm{ROUGE-2}_{\textrm{R}}=\frac{2}{4}=0.5$ and $\textrm{ROUGE-2}_{\textrm{P}}=\frac{2}{3}=0.67$, respectively. ROUGE-L identifies the longest common subsequence $n_\textrm{L}$ of tokens in the same order but not necessarily consecutive. For example, a candidate sentence ``John really loves data science and studies it extensively'' and a reference sentence ``\textit{John} very much \textit{loves data science and} enjoys \textit{it} a lot'' produces $n_{\textrm{L}}=6$, generating recall and precision of $\textrm{ROUGE-L}_{\textrm{R}} = \frac{6}{11}=0.55$ and $\textrm{ROUGE-L}_{\textrm{P}} = \frac{6}{9}=0.67$, respectively.

\subsection{State of the Art Abstractive Summarizers}

Papers With Code maintains a list of state-of-the-art abstractive summarizers tested on the CNN/Daily Mail dataset\footnote{\url{https://paperswithcode.com/sota/abstractive-text-summarization-on-cnn-daily}}. Currently, the top four systems are versions of BART, BRIO, PEGASUS, and MoCa. Since we test our approach against these methods, we provide brief overviews of each system. BART is a ``denoising'' transformer model trained to convert noisy or corrupted text into denoised, uncorrupted text using various permutations of a target sentence (\textit{e.g.}, term masking, deletion, or permutation) \cite{lew:2020}. BART's initial language understanding model is then fine-tuned to perform NLP tasks like abstractive summarization. BRIO splits the summary generation process into two stages: generation using cross-entropy loss and evaluation using contrastive loss \cite{liu2022brio}. Combining these metrics balances probabilities across the summary-to-date during training, producing high-quality summaries even when their maximum-likelihood estimation, the standard scoring method during summary generation, would not have recommended them. An extension of PEGASUS, known as \textit{sequence likelihood calibration} (SLiC), calibrates model-generated sequences with reference sequences in the model's latent space. Calibration refers to the ability to compare the quality of different potential summaries. Rather than apply heuristics to perform this, the authors propose to align candidate sequence likelihoods to the target sequence using a model's latent states during the decode stage of the seq2seq process. MoCa addresses the issue of \textit{exposure bias} during inference (\textit{i.e.}, the trainer only has access to previously predicted tokens rather than ground-truth tokens during search) \cite{zha:2022c}. To correct this, MoCa uses a combination of a generator model and an online model to slowly evolve samples that align generator model scores with online model scores using ranking loss. This is done by modifying the generator's parameters using a \textit{momentum coefficient} based on ranking loss during back-propagation.

\subsection{Sentiment Analysis}

Sentiment analysis is an active research area in NLP, information retrieval (IR), and machine learning (ML). Two standard analysis methods are: (1) supervised, using a training set to build emotion estimation models, and (2) unsupervised, where raw text is converted directly into scores along emotional dimensions
\cite{liu:2012, moh:2015, pang:2008, zha:2018}.

\begin{figure}[!t]
\centering
\hspace*{\fill}
\subfloat[]{\includegraphics[width=0.375\columnwidth]{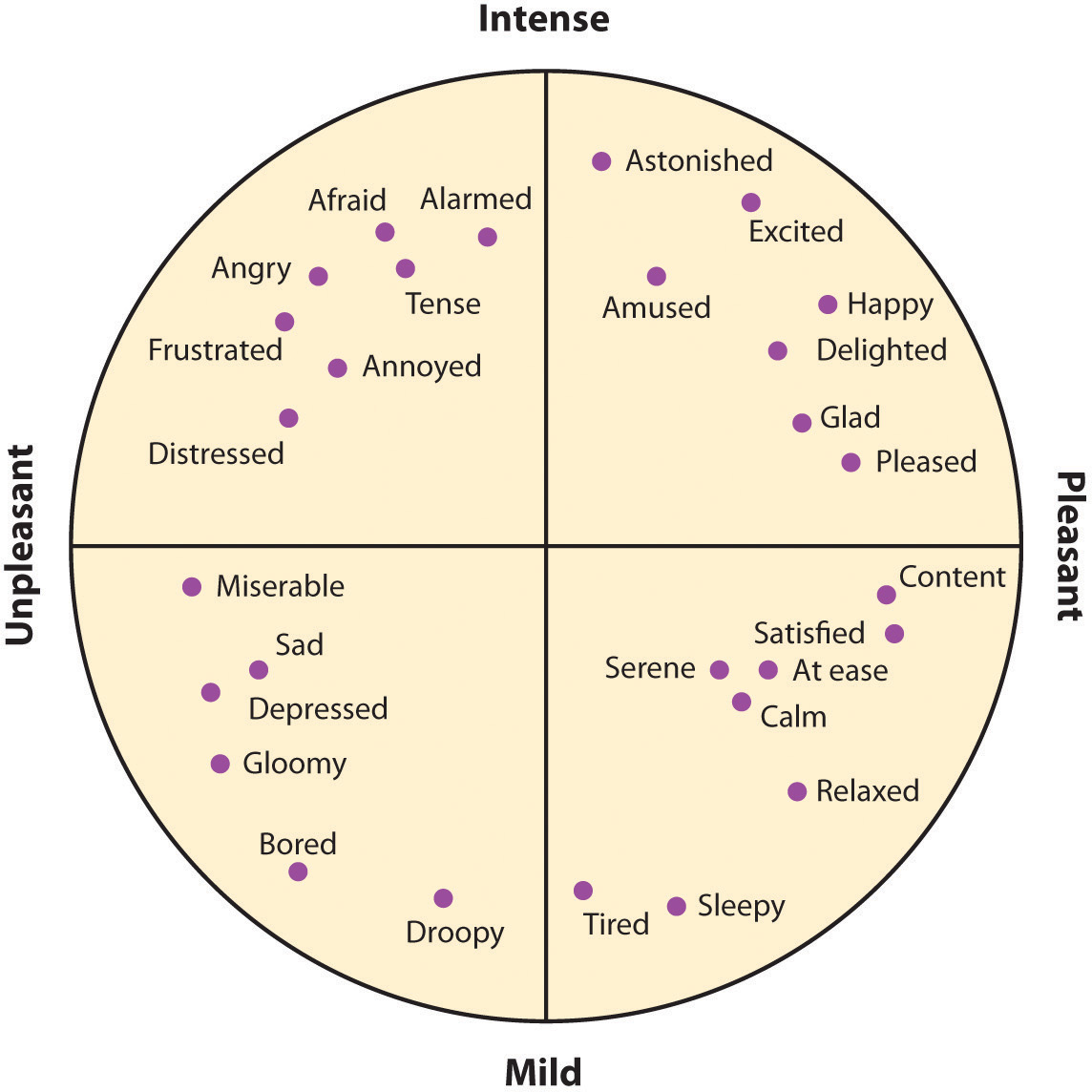}}\hfill%
\subfloat[]{\includegraphics[width=0.375\columnwidth]{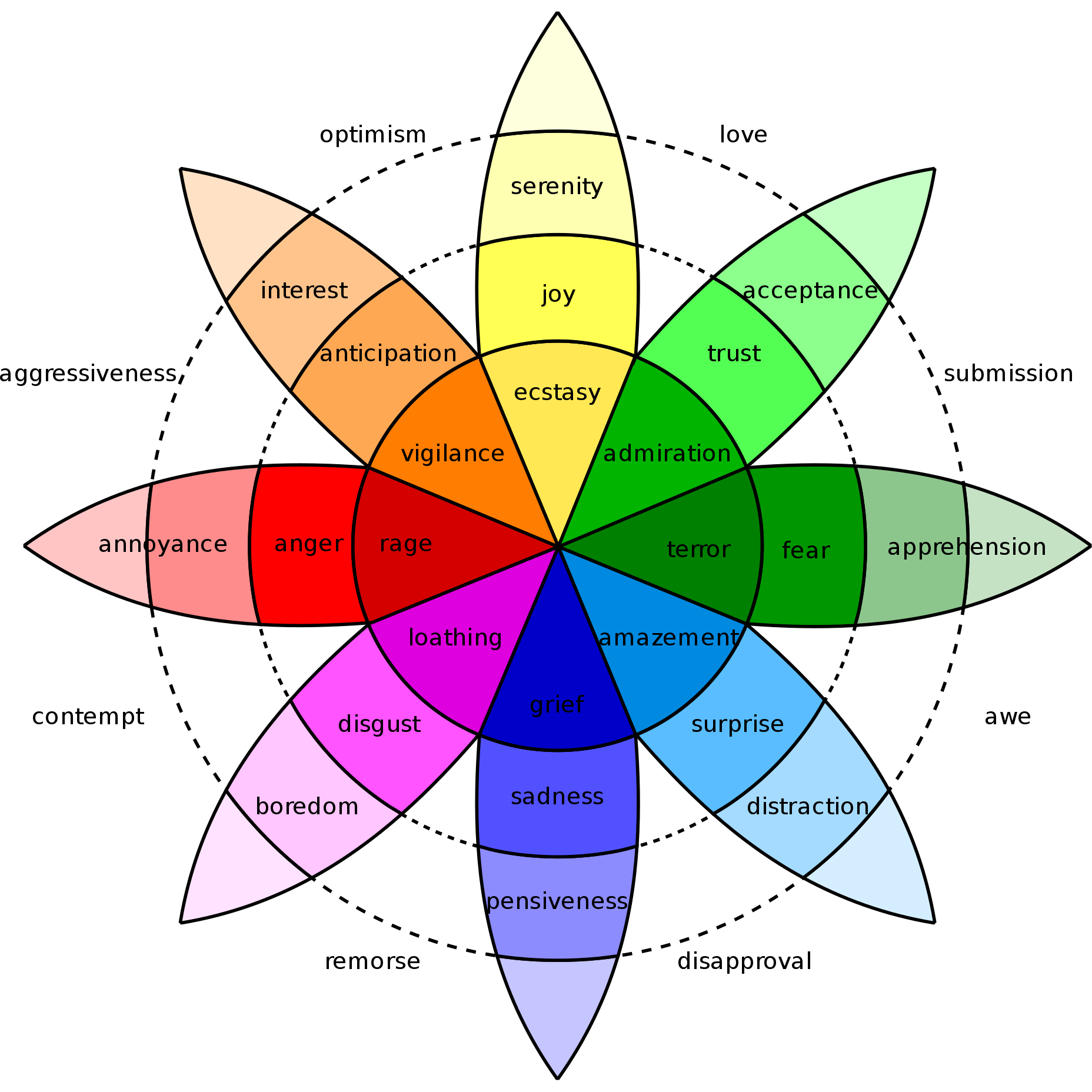}}
\hspace*{\fill}%
\caption{Emotional models: (a) Russell's emotional circumplex, pleasure (valence) on the horizontal axis, arousal on the vertical axis; (b) Plutchik's emotional model, anger--fear on the horizontal axis, joy--sadness on the vertical axis, trust--disgust on the right-diagonal axis, and anticipation--surprise on the left-diagonal axis}
%
\label{emotion-model}
\end{figure}

Analysis is often built on psychological models of emotion that use orthogonal dimensions to describe emotional affect. Russell defined three dimensions pleasure (or valence), arousal, and dominance---the PAD model---to represent emotion \cite{Rus:80, Rus:99} (Figure~\ref{emotion-model}a). Plutchik's four-dimensional model of joy--sadness, anger--fear, trust--disgust, and anticipation--surprise uses a color wheel to represent basic emotions: hue for dimension endpoints (eight hues) and saturation for emotional intensity (weak saturation for low intensity to strong for high, Figure~\ref{emotion-model}b) \cite{plu:80}.

\subsection{Sentiment Estimation}

In supervised NLP approaches, preprocessing has been applied prior to sentiment analysis. Using ML, Pang and Lee calculated subjectivity weights for sentences, producing a graph of sentence nodes and subjectivity-weighted edges \cite{pang:2004}. A minimum graph cut is used to separate objective and subjective sentences. Pang et al. then compared Na\"{i}ve Bayes, maximum entropy, and support vector machines (SVMs) for classifying movie reviews as positive or negative \cite{Pang:2002}. Unigrams performed best using SVMs. Augmenting the training set with intuitive extensions like bigrams, term frequencies, part of speech tagging, and document position information did not improve performance. Turney rated online reviews as positive or negative using pointwise mutual information to generate statistical dependence between review phrases and the anchor words ``excellent'' and ``poor'' \cite{turney:2002}.

Several pre-built sentiment analysis libraries are available \cite{bon:2019,DiB:2021}. In Python, the Natural Language Toolkit's Valence Aware Dictionary and Sentiment Reasoner (NLTK VADER) scores text blocks for polarity (negative, neutral, and positive) and overall sentiment (compound). Textblob includes a sentiment analysis engine, among other common NLP algorithms, returning a sentiment polarity score in the range $[ -1, \ldots, 1 ]$ and a subjectivity score in the range $[ 0, 1 ]$. Flair uses a pre-trained word embedding model to perform sentiment analysis. Although slower than VADER or Textblob, tests suggest that Flair produces more accurate sentiment scores when compared to star ratings for product reviews \cite{pod:2020}.

More recently, deep learning has been applied to sentiment analysis with great success. Initial work focused on recurrent neural networks, often augmented with long-short term memory (RNNs and LSTMs). Recently, RNN and LSTM models have been superseded by deep learning transformers. Bidirectional encoder representations from transformers (BERT) has been fine-tuned for sentiment analysis \cite{kenton:2019}. A common approach is to use TensorFlow and a review database with star ratings like IMDB or Amazon to predict sentiment polarity. Once extended, BERT can be applied to unlabeled text to estimate sentiment. GPT-3's pre-trained model can also be extended to estimate sentiment, although this can often be done with fewer training samples based on its few-shot context abilities. Masked sequence to sequence (MASS) and BART combine the encode--decode step to produce generalizations of BERT and GPT \cite{lew:2020, song_k:2019}. MASS masks out $k$ consecutive tokens in the input sequence, then attempts to predict those tokens in the output sequence. BART introduces noise into the input sequence
to generate ``noisy'' input for the encoder, then applies an autoregressive decoder to remove the noise and reconstruct the original input. Since MASS and BART are extensions of BERT and GPT-3, they can also be extended to estimate sentiment.

\subsubsection{Sentiment Dictionaries.}

A common unsupervised approach employs sentiment dictionaries. Terms appear as keys, each associated with one or more emotional dimension scores.
POMS-ex (Profile of Mood States) is a 793-term dictionary designed to measure emotion on six dimensions: tension--anxiety, depression-dejection, anger--hostility, fatigue--inertia, vigor--activity, and confusion--bewilderment \cite{Pepe:2008}.
ANEW (Affective Norms for English Words) used the PAD model to score 1,033 emotion-carrying terms along each dimension using a nine-point scale \cite{mislove:2010}.
Mohammad and Turney created EmoLex from 14,182 nouns, verbs, adjectives, and adverbs using Plutchik's four emotional dimensions \cite{Moh:2013}.
Other dictionaries also exist: SentiStrength, built from MySpace comments \cite{Thelwall:2010}; LWIC (Linguistic Inquiry and Word Count), a dictionary that classifies terms as positive, negative, or neutral \cite{Tausczik:2010}; and SentiWordNet, built from the well know WordNet synset dictionary \cite{Baccianella:2010}.
More recently, researchers have applied Amazon Mechanical Turk to assign scores for emotional dimensions to large dictionaries. Warriner extended the original ANEW dictionary to approximately 13,000 terms \cite{War:2013} using MTurk to obtain PAD scores and compare results to the original ANEW scores for validation.

\subsubsection{Sentiment Visualization.}
Visualizing sentiment has received significant attention as part of the general text visualization area. Kucher et al. provide an overview of recent sentiment visualization techniques \cite{Kuc:2017}.
Cao et al. developed Whisper to monitor the spatiotemporal diffusion of social media information. Sentiment polarity was visualized using a sunflower metaphor \cite{Cao:2012}.
SocialHelix followed, visualizing and tracking social media topics as they form and their sentiment diverges using a DNA-like representation \cite{Cao:2014}.
Wu et al. presented opinion propagation in Twitter using a combination of streamgraphs and Sankey graphs \cite{Wu:2014}.
Liu et al. linked primary and secondary text using semantic lexical matching. Results are presented in a dashboard containing topic keywords, concept clusters, and a causality timeline \cite{Liu:2017}.
El-Assadi et al. visualized multi-party conversation behavior at the topic level with ConToVi \cite{ElA:2016}. They also extracted conversation threads from large online conversation spaces using a combination of supervised and unsupervised machine learning algorithms \cite{ElA:2018}. Hoque and Carenini implemented ConVis and MultiConVis, an ML, NLP, and visual analytic system to explore blog conversations \cite{Hoq:2014, Hoq:2016}.
Mohammad et al. extracted stance and sentiment in tweets using a labeled database, with results visualized using treemaps, bar graphs, and heatmaps \cite{Moh:2017}.
Kucher et al. identified stance and sentiment polarity in social media text, then used similarity over these properties to visualize analysis of collections of topic--data source streams \cite{Kuc:2020}. Wei et al. proposed TIARA, a system to extract topics that are visualized in an annotated streamgraph \cite{Wei:2010}. D\"{o}rk et al. used Topic Streams, a streamgraph approach to monitoring topics in a large online conversation environment over time \cite{Dor:2010}.

Despite this significant progress, numerous challenges in sentiment estimation continue to exist: more subtle text cues (\textit{e.g.}, sarcasm, irony, humor, or metaphors), a writer's emotion versus what they write (\textit{e.g.}, an author evoking a particular emotional affect), emotion towards different aspects of an entity, stance (\textit{i.e.}, the opinion on a topic), and cross-cultural and domain differences (\textit{e.g.}, ``alcohol'' can be evaluated differently depending on the underlying culture) \cite{moh:2015, moh:2016, pang:2008}.

\section{Methodology}

Although LLMs like GPT can perform few-shot abstractive summarization, they are limited in the input size they support. GPT currently has a 4,096 token maximum, which is too small for even a moderately sized document or document collection. The issue, therefore, becomes: how can we scale an LLM to perform abstractive summarization? An obvious approach is to compress the collection to the LLMs's size limits intelligently. This shifts our goal from abstractive summarization to intelligent document compression prior to summarization. Ideally, we would like to identify, extract, and compress the most semantically meaningful information from the collection. This explains the standard recommendation of extractive summarization followed by abstractive summarization for large document collections. Although existing extractive summarization techniques can produce acceptable results, this reduces the abstractive summarizer to a language rewriter. The choppy and often discontinuous extractive sentences are converted into a grammatically and syntactically correct summary, much more like what a human writer would produce. The main point, however, is that if a concept is not included in the extractive summary, it can never occur in the abstractive summary. Therefore, it is critical to produce the best extractive component or components to summarize.

Overall, the key objectives of our proposed models are: (1) to overcome the term limitation of GPT for the abstractive summarization task and (2) to create an end-to-end abstractive summary generation and visualization pipeline for large document collections. Unlike traditional extractive summary approaches that produce a single set of sentences, we propose to subdivide the document collection into semantic topic clusters, extract representative terms from each cluster that subdivide sentences into semantic chunks, abstractively summarize each semantic chunk, then concatenate the chunks to generate a final abstractive summary. This approach has a number of novel advantages: (1) improving scaling by subdividing a document collection into semantically similar clusters; (2) operating at a semantic chunk level rather than a sentence level; and (3) attempting to identify the most semantically meaningful information in a collection. Once the abstractive summary is produced, sentiment analysis is used to augment it with estimated emotional affect, presented using visualization. Sentiment analysis presents unique challenges, particularly how to aggregate sentiment over a multi-sentence text block and visually represent sentiment and its related properties optimally.

An overview of our approach is as follows. Given a document collection, we first perform text clustering and topic modeling on the collection. We then identify topic-representative terms and construct representative term sets for each cluster, containing a representative term and all semantically similar terms in the parent cluster. To further reduce the topic cluster size, we divide sentences containing representative terms in each cluster into semantic chunks based on change points in their content. We leverage GPT's summarization API to summarize each semantic chunk and use its concatenation API to combine the semantic chunk summaries into an abstractive summarization of the original document collection. Finally, we estimate sentiment and present the summary and its sentiment using an interactive visualization dashboard.

We demonstrate our framework using five 100-document collections collected from the CNN/Daily Mail and Gigaword datasets. These collections cover the topics Barack Obama, university research, wildlife protection, the stock market, and basketball. We compare our results to current state-of-the-art abstractive summarizers to test our framework.

\subsection{Query Support}

If required, query support can be performed prior to our abstractive summary pipeline. Here, we retrieve documents $d_i$ in document collection $D, \; 1 \leq i \leq n_D$ with the highest similarity to a user query $q$. We employ a text encoder $E$ that maps its input to a final hidden layer in an LLM. $E$ is applied to all documents in $D$ and $q$, producing $E_{d_i}, \; 1 \leq i \leq n_D$ and $E_q$. We use Facebook's FAISS library to index all $d_i$. This allows us to query the $u$-nearest matches to $q$ in a computationally efficient manner, generating a subset $D^{\prime}$ of the original $D$, $D^{\prime} = \{ d^{\prime}_1, d^{\prime}_2, \ldots d^{\prime}_u \}$. $D^{\prime}$ replaces $D$ as input to the document clustering stage.

\subsection{Document Clustering}

We apply document clustering to $D$ (the original document collection or the result of a user query) to subdivide $D$ into more granular topic sets. We use the output of a text encoder $E$ together with UMAP (Uniform Manifold Approximation and Projection) to perform projection, allowing us to cluster in a lower dimension like a plane (2D) or volume (3D) \cite{mcinnes2018umap}. Various dimensional reduction approaches are available, including PCA (principal component analysis) and t-SNE (t-distributed Stochastic Neighbor Embedding) \cite{her:2010,maa:2008}. We selected UMAP based on its nonlinear and unsupervised nature and ability to efficiently manage large text datasets, preserving their local and global structure.

Following projection, we apply HDBSCAN to generate topic clusters \cite{cam:2013}. HDBSCAN transforms points into low-density and high-density spatial regions based on local neighbor distances, builds a minimum spanning tree on the resulting distance-weighted graph, constructs and condenses a cluster hierarchy based on a minimum cluster size, then extracts stable clusters from the condensed tree. The result is a set $\mathcal{C}$ of $n_{\mathcal{C}}$ clusters $C_j \in \mathcal{C}, \; 1 \leq j \leq n_{\mathcal{C}}$ that are not constrained to specific shapes or sizes, as well as ``outlier'' or noise documents that do not belong to any cluster.

\setlist[description]{font=\normalfont}

\begin{table*}[!t]
\caption{Example abstractive summarization of a topic cluster discussing ChatGPT}
\label{cluster-summarize}
\centering
\small{
\begin{tblr}{
  colspec={c|X},
  rowspec={Q[m]Q[m]Q[m]Q[m]Q[m]Q[m]Q[m]Q[m]},
  row{odd}={gray9}, row{even}={blue9}
}
{
\textbf{Topic}\\
\textbf{Documents}
} &
{
\begin{minipage}{\hsize}
\vspace{1ex}
\begin{enumerate}[nosep,leftmargin=0.2in]
\item Since it launched in late 2022, ChatGPT seems to have taken the world by storm. $\cdots$ So while it's impossible to predict what a future filled with A.I. lobbyists will look like, it will probably make the already influential and powerful even more so.
\item[] \hspace*{2.2in}$\cdots$
\item[$n$.] Since its launch in November 2022, ChatGPT (`GPT' stands for Generative Pre-trained Transformer), a type of artificial intelligence model, has gained over a million users. $\cdots$ I recommend we do all we can as educators to cultivate the powers of the human mind in the face of this novel threat to our intelligence.
\end{enumerate}
\vspace{1ex}
\end{minipage}
}\\
{
\textbf{LDA}\\
\textbf{Concepts}
} &
{
\begin{minipage}{\hsize}
\vspace{1ex}
\begin{enumerate}[nosep,leftmargin=0.2in]
\item dunn ($w=0.0009$), said  ($w=0.0009$), $\ldots$ lobbi  ($w=0.0009$)
\item[] \hspace*{2.2in}$\cdots$
\item[$n$.] lobbi ($w=0.0033$), comment ($w=0.0030$), \ldots strategi ($w=0.0019$)
\end{enumerate}
\vspace{1ex}
\end{minipage}
}\\
{
\textbf{Topic}\\
\textbf{Term}\\
\textbf{Sets}
} &
{
\begin{minipage}{\hsize}
\vspace{1ex}
\begin{enumerate}[nosep,leftmargin=0.2in]
\item
  \begin{description}[labelwidth=7.67pt,labelsep=0pt,leftmargin=7.67pt]
  \item[\{~] \textbf{lobbi:} $\varnothing$~\}
  \end{description}
\item[] \hspace*{2.2in}$\cdots$
\item[$n$.]
  \begin{description}[labelwidth=7.67pt,labelsep=0pt,leftmargin=7.67pt]
  \item[\{~] \textbf{said:} aforesaid, allege, articulate, aver, enjoin, enounce, enunciate, order, pronounce, read, said, say, sound\_out, state, suppose, tell~\}
  \end{description}
\end{enumerate}
\vspace{1ex}
\end{minipage}
}\\
\textbf{$Sen_{C_j}$} &
{
\begin{minipage}{\hsize}
\vspace{1ex}
\begin{enumerate}[nosep,leftmargin=0.2in]
\item Human lobbyists rely on decades of experience to find strategic solutions to achieve a policy outcome.
\item[] \hspace*{2.2in}$\cdots$
\end{enumerate}
\vspace{1ex}
\end{minipage}
}\\
{
\textbf{Adjacent}\\
\textbf{Sentence}\\
\textbf{Similarity}
} &
{
\begin{minipage}{\hsize}
\centering
\vspace{1ex}
\includegraphics[width=5.5in]{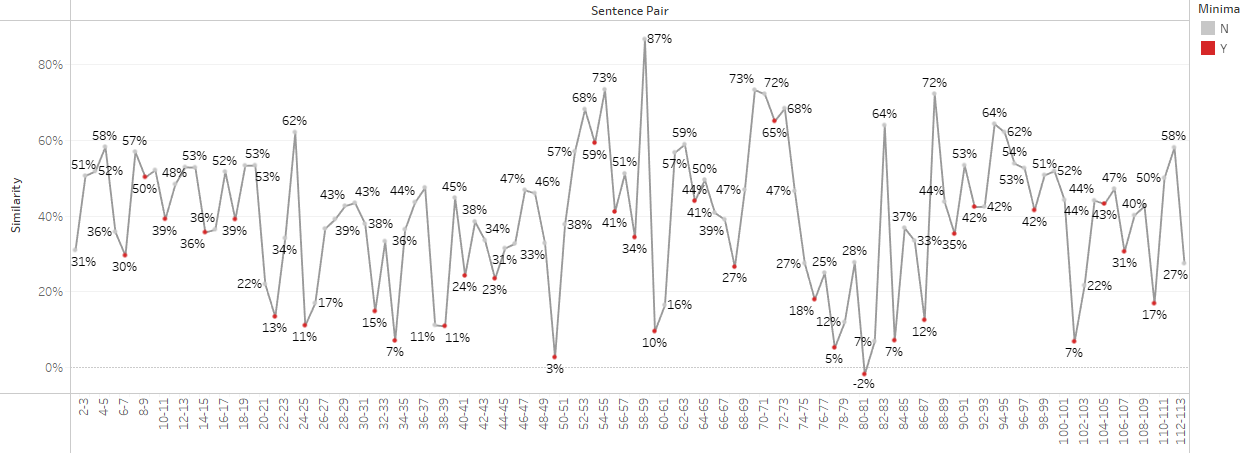}
\vspace{1ex}
\end{minipage}
}\\
{
\textbf{Semantic}\\
\textbf{Chunks}
} &
{
\begin{minipage}{\hsize}
\vspace{1ex}
\begin{enumerate}[nosep,leftmargin=0.2in]
\item Perhaps it could even calibrate the size of donation needed to influence that organization or direct targeted online advertisements carrying a strategic message to its members. Human lobbyists rely on decades of experience to find strategic solutions to achieve a policy outcome. Imagine an A.I.-assisted lobbying firm that can attempt to place legislation in every single bill moving in the U.S. Congress, or even across all state legislatures.
\end{enumerate}
\vspace{1ex}
\end{minipage}
}\\
{
\textbf{Chunk}\\
\textbf{Summaries}
} &
{
\begin{minipage}{\hsize}
\vspace{1ex}
\begin{enumerate}[nosep,leftmargin=0.2in]
\item Human lobbyists can target individuals directly to influence policymaking, and A.I. can be used to understand and target actors within a network.
\item[] \hspace*{2.2in}$\cdots$
\end{enumerate}
\vspace{1ex}
\end{minipage}
}\\
{
\textbf{Abstractive}\\
\textbf{Summary}
} &
\begin{minipage}{\hsize}
\vspace{1ex}
ChatGPT is a chatbot created by OpenAI which has the potential to revolutionize communication, but also raises ethical concerns. Professor Dunn believes it should not be used to generate entire essays, while Professor Porter sees potential in using it to generate ideas. GPT detectors are being developed to protect against GPT, and we need to prioritize relationship and dialogue skills in the classroom and develop courses for working with GPTs and AI text generators.
\vspace{1ex}
\end{minipage}
\end{tblr}
}
\end{table*}

\subsection{Topic Sentence Extraction}

To identify concept-based topic keywords, latent Dirichlet allocation (LDA) is applied to the set of documents in each cluster \cite{ble:2003}. LDA converts bag-of-words-based term-document matrices into concept-document matrices, where each document $d_i \in C_j$ is transformed from a weighted term frequency vector to a concept vector representing the amount of each \textit{latent} concept in cluster $C_j$ that $d_i$ contains.

The first step in LDA is identifying $n_{C_j}$ concepts as weighted combinations of the unique terms contained in $C_j$. $n_{C_j}$ must be defined before running LDA and is itself an open problem. Given $n_{C_j}$, a Dirichlet distribution is assumed to form conditional probabilities of document-concept mixtures and term-topic assignments. This is used to generate a document-concept matrix and a concept-term matrix. The concept-term matrix defines the weights of all unique terms in $C_j$ for each of the $n_{C_j}$ concepts. This term list is usually truncated to contain the top $t_{C_j}$ terms or terms whose weights exceed a predefined threshold $\varepsilon_{C_j}$. Now, every $d_i \in C_j$ is represented by the amount of each $n_{C_j}$ concept $d_i$ contains. This allows similarity to be defined by concept overlap, a more semantic approach than weighted term overlap.

We use the concept-term matrix to construct a collection of \textit{topic terms} $T_{C_j}$. $T_{C_j}$ is the set of terms for $C_j$'s concepts that occur above a threshold frequency across all concepts. $T_{C_j}$ is extended using WordNet to add synonyms for each topic term. We extract all sentences from each $d_i \in C_j$ that contain one or more topic terms, generating a list of topic sentences $\textit{Sen}_{C_j} = \{ \textit{sen}_{1,j}, \textit{sen}_{2,j}, \ldots \}$.

\subsection{Semantic Chunking}

Since the number of sentences in $\textit{Sen}_{C_j}$ may still be large, we split sentences $\textit{sen}_{i,j} \in \textit{Sen}_{C_j}$ into \textit{semantic chunks} $K_{C_j} = \{ k_{1,j}, k_{2,j}, \ldots \}$. To do this, we obtain the SentenceBERT \cite{reimers2019sentencebert} embeddings for each $\textit{sen}_{i,j}$ and use them to construct a similarity matrix $\accentset{\textstyle\sim}{\textit{Sen}_{C_j}}$.
To automatically identify chunking points between sentences, we take the two adjacent diagonals in $\accentset{\textstyle\sim}{\textit{Sen}_{C_j}}$ immediately to the right of the main diagonal and build a two-column matrix \cite{chunk}. The columns represent similarity scores between all pairs of adjacent sentences. To better capture differences in similarity, we amplify the similarity scores for certain sentences and suppress the scores for others using an activation weight $w$ based on the reverse sigmoid function.

\begin{equation}
w(x) = \frac{1}{1 + \exp(0.5 x)}
\label{activation-weight}
\end{equation}

\noindent
where $x$ is the similarity score in each matrix cell. The weighted similarities in each row are summed to compute the similarity between pairs of adjacent sentences. Next, \textit{relative minima} are identified: locations in the weighted sum list where the similarity score decreases, then increases. The relative minima represent the splits between semantic chunks.

\subsection{GPT Zero Shot Summarization}
Given each chunk $k_{i,j} \in K_{C_j}$, we run the summarization pipeline through GPT's completion API

\begin{equation}
\textit{sum}_{i,j} = \textrm{GPT3Summarization}(k_{i,j} + \textrm{Tl;dr:})
\end{equation}

\begin{figure*}[!t]
\captionsetup[subfigure]{justification=centering}
\centering
\subfloat[]{\includegraphics[width=0.95\linewidth]{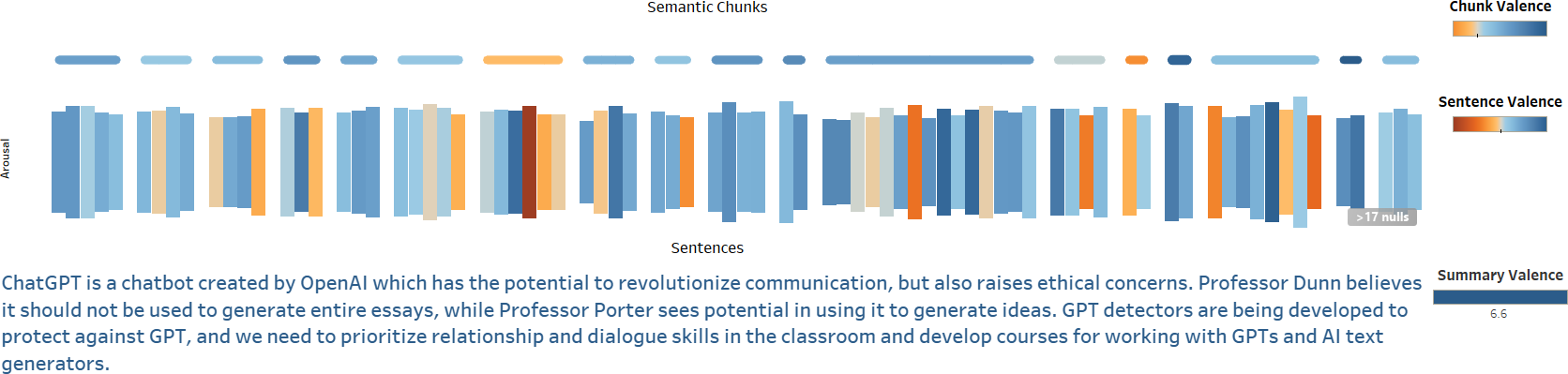}}\\[6pt]%
\subfloat[]{\includegraphics[width=0.95\linewidth]{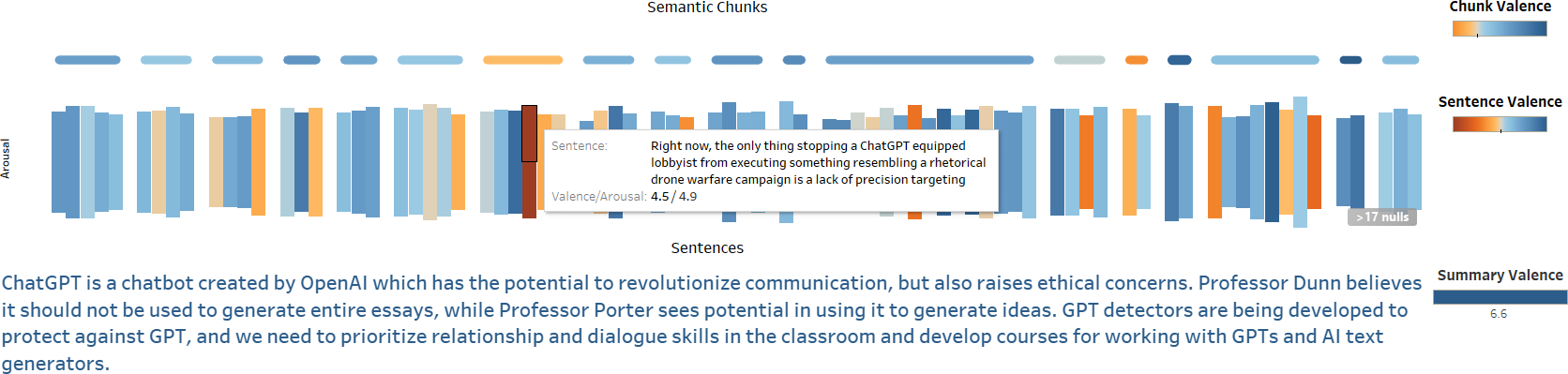}}\\[6pt]%
\subfloat[]{\includegraphics[width=0.95\linewidth]{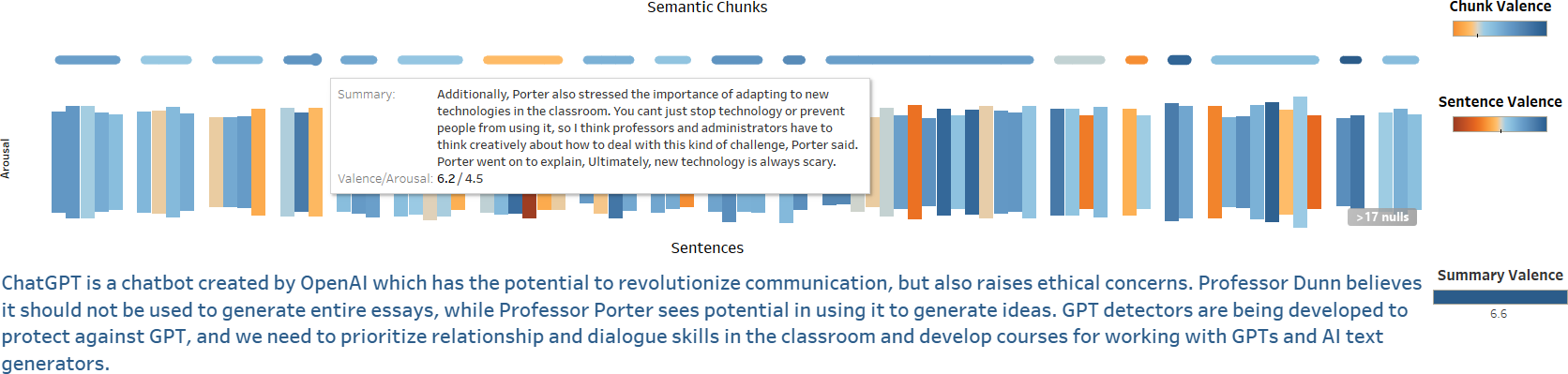}}\\
\caption{Summary visualizations: (a) chunk summary valence (top, represented by hue) and individual sentence valence and arousal (bottom, represented by hue and height) for the ChatGPT topic cluster, the final abstractive summary of the cluster's text is shown at the bottom of the figure; (b) highlighting a sentence to see its text, valence, and arousal; (c) highlighting a semantic chunk to see its text, valence, and arousal}
\label{summary-viz}
\end{figure*}

\noindent
This generates summaries for each chunk. Although we did not encounter this during our analysis and testing, it is \textit{possible} that $k_{i,j}$ exceeds GPT's maximum token count. If this happens, $k_{i,j}$ is further subdivided into two parts. This is done at the internal sentence in $k_{i,j}$ that produces the largest absolute difference with its neighbours based on weighted semantic sentence similarity $\textrm{sim}_{i,j}$. Assuming $k_{i,j}$ spans sentences $k_{i,j} = ( sen_{i,u}, sen_{i,u+1}, \ldots sen_{i,v})$, the split point $sp$ occurs as follows.

\begin{align}
sp &= \max_{u+1 \, \leq \, s \, \leq \, v-1}
\{ \, ||( \textrm{sim}_{i,s-1} - \textrm{sim}_{i,s} )|| + ||( \textrm{sim}_{i,s} - \textrm{sim}_{i,s+1} )|| \, \}
\end{align}

\noindent
producing two semantic chunks $k^{\prime}_{i,sp} = ( sen_{i,u}, \ldots sen_{i,sp} )$ and $k^{\prime}_{i,sp+1} = ( sen_{i,sp+1}, \ldots sen_{i,v} )$ that span the original $k_{i,j}$. Identical subdivision of $k^{\prime}_{sp}$ and $k^{\prime}_{sp+1}$ can occur as needed. Once each semantic chunk meets the maximum term limit constraint, the $\textrm{sum}_{i,j}$ are combined and summarized.

\begin{equation}
\textit{Sum}_j = \textrm{GPT3Summarization}(sum_{1,j} + sum_{2,j} + \cdots )
\end{equation}

\noindent
This produces a final abstractive summary for cluster $C_j$. We repeat the same procedure for all clusters in $D$ to construct a summary for every cluster in the document collection. Finally, we combine cluster summaries, again using GPT.

\begin{equation}
\textit{Sum} = \textrm{GPT3Summarization}( {\textit{Sum}_1 + \textit{Sum}_2 + \cdots } )
\end{equation}

\section{Sentiment Visualization}

To further highlight patterns and insights in the abstractive summaries and the components used to construct them, we designed a sentiment visualization dashboard to allow viewers to explore different aspects of the final summaries. Fig.~\ref{summary-viz} shows an example of the visualization for our ChatGPT topic cluster, including the base visualization (Fig.~\ref{summary-viz}a) and interactive exploration of the raw sentences (Fig.~\ref{summary-viz}b) and semantic chunks (Fig.~\ref{summary-viz}c).

Our approach breaks the abstractive summary into three parts: semantic chunks represented as lines colored by valence on the top; individual sentences shown as bars within each semantic chunk together with their valence (represented by hue) and arousal (represented by height); and the final abstractive summary for the entire topic shown in text colored by valence on the bottom. We apply our knowledge of cognitive vision in visualization to choose perceptually effective representations for our data. For example, we and others have studied extensively the best use of geometric shape, properties of color (luminance, hue, and saturation), layout, and their combined to highlight data properties that we believe are most relevant to our viewers \cite{Cal:89,Hea_TVCG:99,Hea_TVCG:2012}. These guidelines were used to select the double-ended saturation scales, rectangles, and lines to present sentiment, arousal, sentences, and semantic chunks.

\begin{table*}[!t]
\caption{Example queries, keywords (in order of importance) for the LDA topics built from the documents returned for each query, and the final abstractive summary of the query's topic summaries}
\label{query-summarize}
\centering
\begin{tblr}{
  columns={valign=m},
  colspec={c|X|X},
  rowspec={Q[c,m]Q[m]Q[c,m]Q[m]},
  row{odd}={gray9}, row{even}={blue9}
}
{
\textbf{Query}&
\textbf{Topics}&
\textbf{Query Summary}
}\\
{
Barack\\Obama
} &
{
\begin{minipage}{\hsize}
\vspace{1ex}
\begin{enumerate}[nosep,leftmargin=0.15in]

\item\adjustbox{valign=t}
{
\begin{minipage}{\dimexpr\hsize-0.25in}
  \begin{description}[labelwidth=\widthof{\{~},labelsep=0pt,leftmargin=\widthof{\{~}]
  \item[\{~] joke, jest, dinner, attend, $\ldots$ poll, survey, Bush, Walker~\}
  \end{description}
\end{minipage}
}
\item\adjustbox{valign=t}
{
\begin{minipage}{\dimexpr\hsize-0.25in}
  \begin{description}[labelwidth=\widthof{\{~},labelsep=0pt,leftmargin=\widthof{\{~}]
  \item[\{~] kany, claim, title, call, $\ldots$ pant, broad, leg, site~\}
  \end{description}
\end{minipage}
}
\item\adjustbox{valign=t}
{
\begin{minipage}{\dimexpr\hsize-0.25in}
  \begin{description}[labelwidth=\widthof{\{~},labelsep=0pt,leftmargin=\widthof{\{~}]
  \item[\{~] victori, white, White, great,  $\ldots$ country, state, nation, suggest~\}
    \end{description}
\end{minipage}
}
\item\adjustbox{valign=t}
{
\begin{minipage}{\dimexpr\hsize-0.25in}
  \begin{description}[labelwidth=\widthof{\{~},labelsep=0pt,leftmargin=\widthof{\{~}]
  \item[\{~] indic, Indic, Indo-Aryan, zeid, $\ldots$ speaker, plan, design, month~\}
  \end{description}
\end{minipage}
}
\item\adjustbox{valign=t}
{
\begin{minipage}{\dimexpr\hsize-0.25in}
  \begin{description}[labelwidth=\widthof{\{~},labelsep=0pt,leftmargin=\widthof{\{~}]
  \item[\{~] obama, cotton, cotton\_fiber, cotton\_wool, $\ldots$ rock, sway, site, user~\}
    \end{description}
\end{minipage}
}
\item\adjustbox{valign=t}
{
\begin{minipage}{\dimexpr\hsize-0.25in}
  \begin{description}[labelwidth=\widthof{\{~},labelsep=0pt,leftmargin=\widthof{\{~}]
  \item[\{~] line, deduct, republican, budget, $\ldots$ lexu, said, say, tell~\}
    \end{description}
\end{minipage}
}
\item\adjustbox{valign=t}
{
\begin{minipage}{\dimexpr\hsize-0.25in}
  \begin{description}[labelwidth=\widthof{\{~},labelsep=0pt,leftmargin=\widthof{\{~}]
  \item[\{~]walk, walking, base\_on\_balls, pass, $\ldots$ auctioneer, Bush, cook, sell~\}
    \end{description}
\end{minipage}
}
\end{enumerate}
\vspace{1ex}
\end{minipage}
} &
{
Scott Walker recently signed a right-to-work law in Wisconsin, and Obama has criticized it as part of a "sustained, coordinated assault on unions." Unions are protesting the law, while proponents say it will bring business investment and create jobs. Walker is campaigning for the Republican presidential primary, and his super PAC is defending his right-to-work law. Jeb Bush is fundraising for his Right to Rise PAC and Rand Paul has criticized George W. Bush for not being conservative enough.
}\\
$\cdots$ & $\cdots$ & $\cdots$ \\
{
wildlife\\protection
} &
{
\begin{minipage}{\hsize}
\vspace{1ex}
\begin{enumerate}[nosep,leftmargin=0.15in]

\item\adjustbox{valign=t}
{
\begin{minipage}{\dimexpr\hsize-0.25in}
  \begin{description}[labelwidth=\widthof{\{~},labelsep=0pt,leftmargin=\widthof{\{~}]
  \item[\{~] duck, dog, water, shoot, $\ldots$ monkey, meat, lamb, dean~\}
  \end{description}
\end{minipage}
}
\item\adjustbox{valign=t}
{
\begin{minipage}{\dimexpr\hsize-0.25in}
  \begin{description}[labelwidth=\widthof{\{~},labelsep=0pt,leftmargin=\widthof{\{~}]
  \item[\{~] tiger, beaver, deer, man, $\ldots$ word, pine, southern, girl~\}
  \end{description}
\end{minipage}
}
\item\adjustbox{valign=t}
{
\begin{minipage}{\dimexpr\hsize-0.25in}
  \begin{description}[labelwidth=\widthof{\{~},labelsep=0pt,leftmargin=\widthof{\{~}]
  \item[\{~] wildebeest, gnu, cub, tourist, impala, bear, carry~\}
    \end{description}
\end{minipage}
}
\item\adjustbox{valign=t}
{
\begin{minipage}{\dimexpr\hsize-0.25in}
  \begin{description}[labelwidth=\widthof{\{~},labelsep=0pt,leftmargin=\widthof{\{~}]
  \item[\{~] plant, carbon, tree, reel, $\ldots$ prey, behaviour, behavior, attack~\}
  \end{description}
\end{minipage}
}
\item\adjustbox{valign=t}
{
\begin{minipage}{\dimexpr\hsize-0.25in}
  \begin{description}[labelwidth=\widthof{\{~},labelsep=0pt,leftmargin=\widthof{\{~}]
  \item[\{~] canal, duct, epithelial\_duct, channel, $\ldots$ alternate, author, boat, hunt~\}
    \end{description}
\end{minipage}
}
\item\adjustbox{valign=t}
{
\begin{minipage}{\dimexpr\hsize-0.25in}
  \begin{description}[labelwidth=\widthof{\{~},labelsep=0pt,leftmargin=\widthof{\{~}]
  \item[\{~] tour, circuit, go, spell, $\ldots$ flash, photograph, approach, cover~\}
    \end{description}
\end{minipage}
}
\item\adjustbox{valign=t}
{
\begin{minipage}{\dimexpr\hsize-0.25in}
  \begin{description}[labelwidth=\widthof{\{~},labelsep=0pt,leftmargin=\widthof{\{~}]
  \item[\{~] heron, Hero, Heron, octopu, $\ldots$ drown, flight, green, know~\}
    \end{description}
\end{minipage}
}
\end{enumerate}
\vspace{1ex}
\end{minipage}
} &
{
Jiri Michal was trying to take a picture of a great grey owl in Vysoka when it started playing peek-a-boo with him. Snowy owls are diurnal and hunt silently, but animal rights campaigners have accused the Harry Potter studio tour of mistreating owls by keeping them in cages and allowing fans to touch them. PETA has called for Warner Brothers to stop using live animals in their tour.
}\\
\end{tblr}
\end{table*}

A common issue with sentiment analysis is aggregation. Normally, text should be divided into blocks that contain a single sentiment. Aggregating multiple sentiment values often leads to neutral results since positive and negative sentiments cancel under most aggregation operations, resulting in overall neutral sentiment. Outlier situations, either in the number of positive or negative sentiment scores or in their absolute values, are required to ``push'' the aggregated sentiment toward a positive or negative overall result. Different approaches can address this, for example, setting specific thresholds to discretize valence into negative, neutral, and positive in ways that better distinguish negative and positive scores.

We use a more straightforward method, subdividing semantic chunks into sentences. It is usually assumed that a sentence contains a single sentiment, so aggregating the valence of terms in the sentence should not cause conflicting valence values to cancel. This is why we present individual sentence valence and arousal bars. Several insights can be drawn from this level of detail. First, very few bars are grey, suggesting that few sentences have a neutral valence. At the next level of detail, sentiment chunk valence, only two lines are negative (orange--red), even though orange and red sentence bars occur throughout the topic's text. This shows how negative valence is cancelled by positive valence in a semantic chunk unless large negative sentence valence occurs (seventh semantic chunk) or the number of sentences is small (fourteenth semantic chunk).

At the highest level of detail, the topic's abstractive summary text is displayed and coloured based on its valence. The dashboard lets users hover over sentence bars and semantic chunk lines interactively to reveal their underlying text, valence, and arousal scores (Fig.~\ref{summary-viz}b,c).

\section{Experiments}

We evaluated our proposed framework for abstractive summarization by comparing it to existing state-of-the-art techniques: BART, BRIO, PEGASUS, and MoCa. We aimed to investigate whether our proposed framework can maintain zero-shot performance comparable to competing systems while simultaneously handling large multi-document collections. We used the ROGUE-1, ROGUE-2, and ROGUE-L scores to measure summary quality.

\subsection{Datasets}

We performed experiments using the CNN/Daily Mail and the Gigaword datasets from the Hugging Face library. Each entry in the CNN/Daily Mail dataset contains a document ID, a newspaper article's text, and a corresponding summary (or highlights, as it is referred to within the dataset). The Gigaword dataset includes an English news article's text and a corresponding summary. The CNN/Daily Mail dataset contains 286,817 training pairs, 13,368 validation pairs, and 11,487 test pairs. The Gigaword dataset contains approximately 3.8 million training pairs, 189,000 validation pairs, and 2,000 text pairs.


We assess the zero-shot performance of our framework for abstractive summarization by evaluating its ability to perform the summarization task on the test set without any prior training. Given the short length of each article, we increase the size of the document collection $D$ to $n_{D}=100$ documents. For our system, we had to predefine the number of topic clusters  $n_{\mathcal{C}}$ created by HDBSCAN. We chose $n_{\mathcal{C}}=10$ to ensure semantic chunks that met GPT-3's maximum token limit. We also selected GPT hyperparameters $\textrm{temperature}=0.3$ to set randomness in the output to favor terms with a higher probability of occurrence, $\textrm{top}_{p}=0.9$ to select the smallest collection of terms whose cumulative probability exceeds $0.9$, frequency and presence penalties of $0$ to reduce the likelihood of repetitive text, and use of the \textsf{davinici-003} model to ``produce higher quality writing with better long-form generation versus \textsf{davinici-002}\footnote{\url{https://help.openai.com/en/articles/6779149-how-do-text-davinci-002-and-text-davinci-003-differ}}.''

\begin{table*}[!t]
\caption{ROGUE scores for BART, BRIO, PEGASUS, MoCa, and our proposed system}
\label{summary-scores}
\centering
\begin{tblr}{
  colspec={l|c c c|c c c},
  rowspec={Q[m]Q[m]Q[m]Q[m]Q[m]Q[m]},
  row{odd}={gray9}, row{even}={blue9},
  row{1,2}={gray7},
  cell{1}{2,5}={c=3}{c}
}
 & \textbf{CNN/Daily Mail} & & & \textbf{Gigaword}\\
{
\textbf{System} &
\textbf{ROGUE-1} & \textbf{ROGUE-2} & \textbf{ROGUE-L} &
\textbf{ROGUE-1} & \textbf{ROGUE-2} & \textbf{ROGUE-L}
}\\
BART &
44.2 & 21.3 & 40.9 &
39.1 & 20.1 & 36.4\\
BRIO &
48.0 & 23.8 & 44.7 &
--- & --- & ---\\
PEGASUS &
44.2 & 21.5 & 41.1 &
39.1 & 19.9 & 36.2\\
MoCa &
48.9 & 24.9 & 45.8 &
\textbf{39.6} & \textbf{20.6} & \textbf{36.8}\\
Ours &
\textbf{58.7} & \textbf{25.6} & \textbf{56.0} &
38.7 & 19.7 & 35.8\\
\end{tblr}
\end{table*}

\subsection{Performance}

We compared our results against four state-of-the-art abstract summarizers: BART, BRIO, PEGASUS, and MoCa. All four systems are designed to summarize individual documents that meet the limitations of GPT-3's token maximum. None of the systems are specifically built to summarize document collections. Because of this, we ensured that documents with lengths at or below GPT-3's limits were selected. These documents averaged approximately 500 terms.

Since we are generating multi-document summaries for our system, we first need to define a ground-truth summary to compare to. This is done by taking the individual ground-truth summaries from the test dataset for each document in a cluster and concatenating them. The abstractive summary for the topic cluster is compared to the concatenated ground-truth summary. ROGUE scores for each topic cluster are averaged to generate an overall ROGUE score for our approach.

We made five queries (Barack Obama, university research, wildlife protection, stock market, and basketball) on the CNN/Daily Mail and Gigaword datasets to extract 100 documents per query. We applied our system to each query's 100 documents: HDBSCAN was used to generate topic clusters, LDA identified ten concepts per cluster, topic sets were built to identify sentences containing topic terms, adjacent sentence similarities were used to locate relative minima separating semantic chunks, each semantic chunk was summarized using GPT-3, the chunk summaries were concatenated with GPT-3 to produce a topic summary, and the topic summaries were themselves concatenated to generate a final abstractive summary of the original 100 documents. Examples of the queries, the LDA topics generated from the documents returned by the query, and the final abstractive summary we built from individual topic summaries are shown in Table~\ref{query-summarize}. Topic summaries were compared to the ground-truth summaries we constructed to calculate ROGUE scores. The topic cluster ROGUE scores were then averaged to produce a final ROGUE score for our abstractive summarization approach.

We compared the ROUGE-1, ROUGE-2, and ROUGE-L metrics for the five abstractive summaries generated by our system to the competing approaches' ROGUE results. Scores in Table~\ref{summary-scores} for the competing systems were taken directly from their papers' reported results. Our system produced the highest ROGUE scores for the CNN/Daily Mail dataset, and MoCa reported the highest ROGUE scores for the Gigaword dataset. Our Gigaword scores, however, are comparable to all four systems. This is promising since we are generating multi-document summaries versus the other systems that generate single-document summaries. We also wanted to compare to GPT since it is capable of multi-document abstractive summarization. Unfortunately, we could not locate reported ROGUE scores for abstractive summarization. We considered testing GPT directly, but GPT-3 and GPT-4 have a limit of 4,096 and 8,192 input terms, respectively, smaller than our smallest cluster's size. A beta version of GPT-4 claims to increase this limit to 32,767 terms, but as of our testing, neither the GPT-4 nor the GPT-4 beta APIs are available outside of special case use, which we could not secure.

To search for statistical performance differences, we ran the competing systems and our system on the sets of documents generated by our five queries, then performed analysis of variance (ANOVA) on the individual ROGUE scores for each ROGUE type and summarization system. The following steps were performed for each system to calculate an overall ROGUE score.

\begin{enumerate}

\item Ask the given system, for example, BART, to provide an abstractive summary $\textit{Sum}_{\textrm{BART}}$ for an input document.

\item Calculate the ROGUE-1, ROGUE-2, and ROGUE-L scores for $\textit{Sum}_{\textrm{BART}}$ versus the ground-truth summary provided in the test dataset.

\item Average the ROGUE-1, ROGUE-2, and ROGUE-L scores to generate an overall ROGUE score for the given system.

\end{enumerate}

Because variance across summarization systems was not equivalent, we applied the non-parametric Kruskal-Wallis ANOVA. All systems except MoCa offer implementations in the Hugging Face transformer model repository\footnote{\url{https://huggingface.co/models}}. Because of this, MoCa was excluded from our statistical analysis. In addition, BRIO's implementation is not tuned for the short documents in the Gigaword dataset, so it was not included in any ANOVAs involving Gigaword comparisons. For an $\alpha=95$\% significance rate, the $F$-results for the CNN/Daily Mail and Gigaword ROGUE-1, ROGUE-2, and ROGUE-L scores are shown in Table~\ref{summary-ANOVA}.

ANOVA results confirmed a significant difference for all ROGUE scores across dataset and technique. Dunn's post hoc analysis was performed to search for pairwise differences in performance. The following pairs were identified as \textit{not} significantly different.

\begin{itemize}

\item CNN/Daily Mail, ROGUE-1, ours--PEGASUS, $p=0.12$
\item CNN/Daily Mail, ROGUE-2, ours--BART, $p=0.31$
\item CNN/Daily Mail, ROGUE-2, ours--PEGASUS, $p=0.14$
\item CNN/Daily Mail, ROGUE-L, ours--BART, $p=0.11$
\item CNN/Daily Mail, ROGUE-L, ours--PEGASUS, $p=0.11$
\item Gigaword, ROGUE-2, ours--BART, $p=0.10$

\end{itemize}

\begin{table*}[!t]
\caption{Analysis of variance on ROGUE scores for our proposed system versus BART, BRIO, and PEGASUS on the CNN/Daily Mail dataset, and BART and PEGASUS on the Gigaword dataset}
\label{summary-ANOVA}
\centering
\begin{tblr}{
  colspec={l|c c c},
  rowspec={Q[m]Q[m]Q[m]},
  row{odd}={gray9}, row{even}={blue9},
  row{1}={gray7}
}
{
\textbf{Dataset} &
\textbf{ROGUE-1} & \textbf{ROGUE-2} & \textbf{ROGUE-L}
}\\
{CNN} &
{$F(3,1542)=86.93$, $p < 0.01$} &
{$F(3,1542)=27.43$, $p < 0.01$} &
{$F(3,1542)=67.46$, $p < 0.01$}\\
Gigaword &
{$F(2,1045)=72.73$, $p < 0.01$} &
{$F(2,1045)=62.62$, $p < 0.01$} &
{$F(2,1045)=72.75$, $p < 0.01$}\\
\end{tblr}
\end{table*}

Dunn's pairwise results show that our method performs statistically equivalently to PEGASUS and BART for all but one of the ROGUE scores on the CNN/Daily Mail dataset and equivalently to BART for the ROGUE-2 scores on the Gigaword dataset. This is a positive indication of the usefulness and generalizability of our technique to scale to multi-document collections that existing systems either struggle with or cannot handle due to maximum input term limits.

Finally, we note that the results for the Gigaword dataset in Tables~\ref{summary-scores} and~\ref{summary-ANOVA} are mainly due to the types of summaries Gigaword provides, coupled with our use of GPT. GPT is designed to return summaries that contain complete, grammatically correct sentences. Gigaword's summaries are often text snippets and therefore do not correspond as well to GPT's summaries as those included in the CNN/Daily Mail dataset. For example, Gigaword's test dataset contains text--summary entries like ``\textsf{UNK}--\textsf{russian liberal party wins resignation}'' or ``\textsf{the rand gained ground against the dollar at the opening here wednesday , to \#.\#\#\#\# \/ \#\# to the greenback from \#.\#\#\#\# \/ \#\# at the close tuesday .}--\textsf{rand gains ground}''. The first example contains no original text, reporting it as \textsf{UNK} or unknown. The second example uses placeholders for numeric values and a short text snippet rather than a complete sentence for the ground-truth summary. In both cases, GPT, and by extension, our system, struggles to produce a comparable summary. We avoided entries with \textsf{UNK} in either the text or the summary. We did not remove pairs with incomplete or grammatically incorrect summaries since we wanted to honestly represent how our system performs on these types of datasets.

\section{Conclusions and Future Work}

Our goal in this paper is a technique that can scale to perform abstractive summarization on a multi-document collection. We divide documents into semantic topic clusters using FAISS and HDBSCAN. Representative term sets are generated for each cluster, then used to reduce the cluster size into semantic chunks. GPT is applied to summarize each chunk, then concatenate the summaries into an abstractive summarization of each topic. The same concatenation operation combines topic summaries into an overall document collection summary. The sentence, semantic chunk, and topic summaries are analyzed for sentiment, then visualized using an interactive dashboard that allows users to explore the valence, arousal, and raw and summarized text at multiple levels of detail.

Statistical analysis of ROGUE scores for our system and competing approaches, including BART, BRIO, and PEGASUS, confirmed comparable performance for our multi-document summaries versus existing approaches' individual document results. We offer the following advantages over existing systems.

\begin{enumerate}

\item The ability to scale to multi-document collections versus individual documents.

\item Identification of topics using semantic clustering to provide multiple levels of detail on the content of a document collection.

\item Perceptually-based, interactive visualization dashboards designed to present text sentiment, text summaries, and raw text at different levels of detail.

\item Harnessing and extending the capabilities of large language models for abstractive summarization.

\item Future integration of new techniques, for example, new large language models, abstractive summarization algorithms, or evaluation methods, as they become available since our approach can quickly generalize to any of these changes.

\end{enumerate}

In terms of future work, we are currently investigating three potential improvements to our system.

\begin{enumerate}

\item \textbf{Streaming.} Extend our system to support real-time streaming, allowing it to dynamically add or remove documents in the document collection. This would impact topic clustering since we assume topics will shift over time. Real-time clustering algorithms exist, for example, Real Time Exponential Filter Clustering (RTEFC) and Real Time Moving Average Clustering (RTMAC). A potentially more relevant approach is density-based clustering for real-time stream data \cite{chen:2007}.  Another possibility is to track estimated error in the current cluster results and perform an updated clustering when a threshold error is crossed, similar to how we maintained accurate TF-IDF scores in a streaming document environment \cite{ven:2010}. Once clusters are defined, follow-on semantic chunking, chunk, topic, and document collection summarization, and visualization would be performed as in the current system.

\item \textbf{Visualization.} Improve the visualization dashboard to support more sophisticated visual analytics. Extending the visualization dashboard to support additional exploratory analysis, particularly at different levels of detail, is an area of potential interest. Our current focus is a system similar to one we built for exploring topics and their associated sentiment patterns during customer chat sessions \cite{Hea_CGF:2021}. This system was specifically designed to present relevant information at multiple levels of detail.

\item \textbf{Alternative LLMs.} Explore the strengths and limitations of additional LLMs like Bard, BLOOM \cite{le_scao:2023}, and LLaMA that offer abstractive summarization and text concatenation capabilities. We plan to investigate these and similar LLMs to determine whether they have any particular strengths or limitations for our summarization pipeline.

\end{enumerate}

\balance
\bibliographystyle{unsrtnat}
\bibliography{paper}  

\end{document}